\documentclass[letterpaper]{article} 
\usepackage{aaai2027}  
\usepackage[hyphens]{url}  
\usepackage{graphicx} 
\usepackage{natbib}  
\usepackage{caption} 
\usepackage{algorithm}
\usepackage{algorithmic}

\usepackage{newfloat}
\usepackage{listings}
\DeclareCaptionStyle{ruled}{labelfont=normalfont,labelsep=colon,strut=off} 
\floatstyle{ruled}
\newfloat{listing}{tb}{lst}{}
\floatname{listing}{Listing}

\usepackage{booktabs}
\usepackage{multirow}
\usepackage{amssymb}
\usepackage{amsmath}

\title{Q-CamTuning: A Multimodal Large Language Model for Fine-Grained Quality Reasoning in Livestreaming Camera Tuning}
\author {
    Yanchao Bi\textsuperscript{\rm 1},
    Pengfei Chen\textsuperscript{\rm 1},
    Yulong Yan\textsuperscript{\rm 2},
    Qian Zhang\textsuperscript{\rm 2},
    Biao Wang\textsuperscript{\rm 2},
    Leida Li\textsuperscript{\rm 1}\corresponding
}
\affiliations {
    \textsuperscript{\rm 1}School of Artificial Intelligence, Xidian University\\
    \textsuperscript{\rm 2}Huawei Technologies Ltd.\\
    25171111423@stu.xidian.edu.cn, chenpengfei@xidian.edu.cn, ldli@xidian.edu.cn,
    ylyan17@fudan.edu.cn,
    qian.zh@huawei.com,
    wang-biao106@huawei.com
}

\begin{document}

\maketitle

\begin{abstract}
With the popularization of livestreaming applications, automatic camera tuning has become indispensable for ordinary users to enjoy high Quality of Experience (QoE). However, the existing Image Quality Assessment (IQA) methods face two challenges in camera tuning. First, they struggle to identify subtle quality difference between images, which is common in camera tuning. Second, the ability to explain fine-grained differences and to provide suggestions for camera tuning are even more lacking. To address these challenges, we first construct FGLive-Comments, an instruction tuning dataset with rich annotations of both preferences for fine-grained image pairs and quality reasoning from global and local visual evidence, which supports selective adjustment recommendations of camera parameters. Based on FGLive-Comments, we develop Q-CamTuning, a multimodal large language model for livestreaming camera tuning. Its progressive perception–evidence–guidance framework learns quality-aware visual representations, enhances fine-grained reasoning with global–local evidence, and maps quality issues to camera-parameter adjustment directions. Extensive experimental results demonstrate that Q-CamTuning outperforms existing IQA models in both quality scoring and fine-grained image pair ranking. User study also validates its ability to recommend camera tuning suggestions.
\end{abstract}


\section{Introduction}

With the rapid growth of mobile livestreaming in entertainment, e-commerce, and content creation, automatic camera tuning has become increasingly important for image quality and user experience~\cite{gilstrap2023mobile,hamilton2014streaming}. Unlike traditional photography, livestreaming does not allow for post-capture correction, i.e. camera settings must be adjusted in real time as scenes and parameter configurations change. A typical tuning system generates candidate images of the scene under different configurations and uses image quality assessment (IQA) to perform parameter selection and refinement~\cite{sheng2026tuningiqa}. Therefore, IQA should not only compare the relative quality of candidate images but also guide subsequent quality improvement.

\begin{figure}[t]
\centering
\includegraphics[width=\columnwidth]{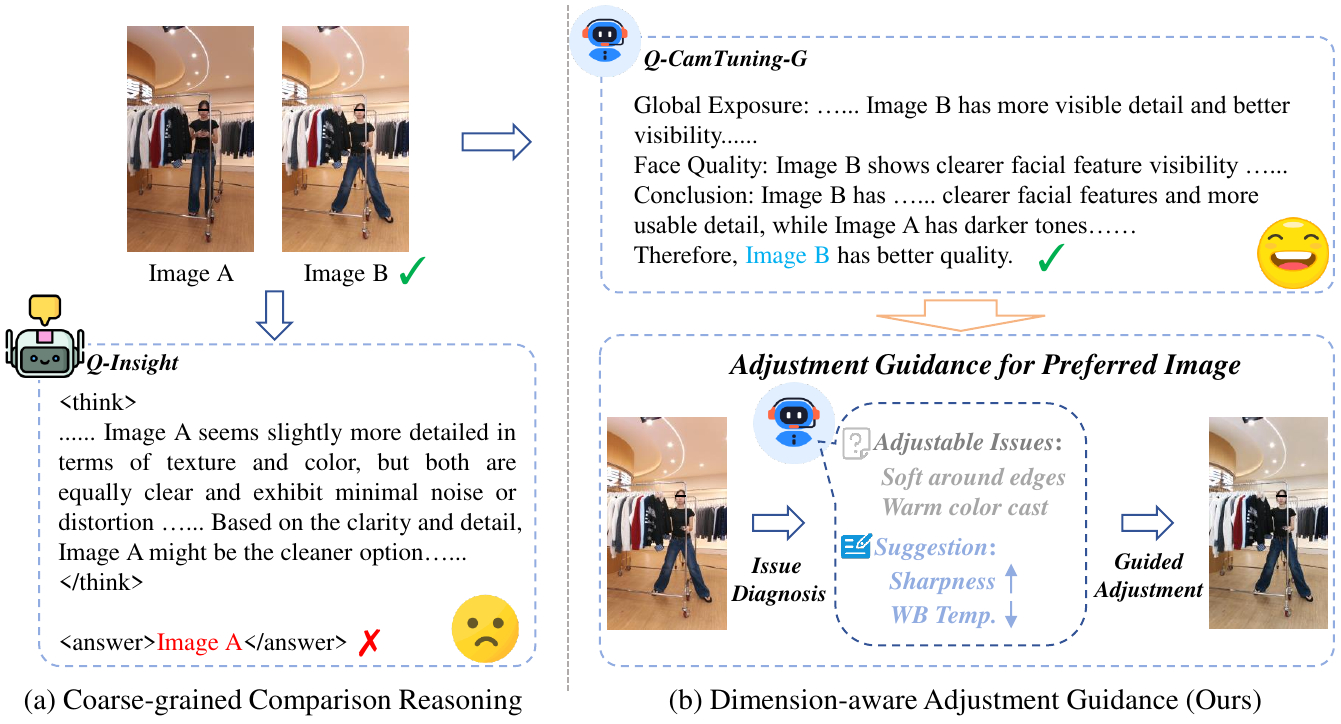} 
\caption{Motivation. Coarse-grained preference reasoning may yield incorrect attribute-aware judgment, while the proposed Q-CamTuning performs comparison and further recommends camera-parameter adjustment directions.}
\label{fig1}
\end{figure}

Most existing blind image quality assessment (BIQA) methods represent image quality using a single scalar score~\cite{mittal2012no,su2020blindly,ke2021musiq,yang2022maniqa}. Although effective for distinguishing pronounced quality differences, they provide limited support for fine-grained camera adjustment. Small parameter variations often produce images with highly similar content whose differences appear only in specific attributes, such as noise, sharpness, exposure, and face quality. To address the above challenge, TuningIQA~\cite{sheng2026tuningiqa} introduced FGLive-10K and jointly learns overall quality regression and pairwise ranking from multi-attribute quality and fine-grained preference annotations. However, it provides only numerical quality predictions and preference results, without explaining the visual evidence underlying its judgments. Such interpretability is essential for camera tuning, where reliable parameter adjustment requires understanding which quality attributes determine the preference.

Multimodal large language models (MLLMs) have extended image quality assessment beyond score prediction to degradation perception and quality reasoning~\cite{wu2024qbench,you2024depicting,wu2024towards}. Existing methods model quality using text-defined levels, pairwise comparisons, or score distributions~\cite{wu2023q,zhu2024adaptive,you2025teaching}, while reinforcement-learning-based approaches further improve degradation perception and quality reasoning~\cite{li2026q}. 

MLLM-based IQA extends quality assessment beyond score prediction to distortion identification, image comparison, and quality reasoning. However, two limitations remain in camera-tuning scenarios. First, these methods may fail to perceive subtle differences between visually similar images, even when producing plausible explanations. Second, they stop at quality assessment and cannot identify actionable issues in the preferred image or recommend camera-parameter adjustment directions. Thus, the process from fine-grained quality perception to tuning guidance remains incomplete.

To address these limitations, we build FGLive-Comments, a multi-modal dataset for camera tuning with human–machine collaborative annotation, and based upon which we further propose Q-CamTuning, a new multimodal large language model for camera tuning quality reasoning. FGLive-Comments provides global and local evidence for perceptible differences in sharpness, noise, exposure, and face quality, together with residual quality issues and selective parameter adjustment suggestions. Q-CamTuning adopts a progressive perception--evidence--guidance framework: it first establishes quality perception through scoring and pairwise ranking, then organizes visual evidence to support fine-grained preference reasoning, and finally maps actionable issues recommendations to camera-parameter adjustment.

The main contributions of this work are summarized as follows:
\begin{itemize}
    \item We construct FGLive-Comments, an instruction-tuning dataset that provides evidence-grounded explanations of fine-grained quality differences and selective parameter guidance for livestreaming camera tuning.
    
    \item We propose Q-CamTuning, a new multimodal large language model that progressively learns quality perception, evidence-based reasoning, and parameter-direction recommendation, unifying fine-grained quality assessment with camera-tuning guidance.
    
    \item We validate Q-CamTuning through quantitative evaluation of quality scoring and ranking, and subjective evaluation of quality reasoning and parameter guidance, demonstrating reliable quality discrimination capability and useful tuning recommendations.
\end{itemize}

\section{Related Work}

\subsection{Blind and Fine-Grained Image Quality Assessment}

BIQA predicts perceptual quality without a reference image. Early methods, such as BRISQUE~\cite{mittal2012no} and NIQE~\cite{mittal2012making}, measured deviations from natural scene statistics. Deep models, including MetaIQA~\cite{zhu2020metaiqa}, HyperIQA~\cite{su2020blindly}, TReS~\cite{golestaneh2022no}, MUSIQ~\cite{ke2021musiq}, and MANIQA~\cite{yang2022maniqa}, improved quality prediction by learning content- and distortion-aware representations. However, most BIQA methods output a single quality score and cannot explicitly distinguish subtle differences between images. Pairwise methods such as RankIQA~\cite{liu2017rankiqa} and PieAPP~\cite{prashnani2018pieapp} address this limitation by directly learning relative quality preferences.

For camera tuning, TuningIQA~\cite{sheng2026tuningiqa} introduced FGLive-10K, which provides overall and multi-attribute quality annotations together with fine-grained pairwise preferences. TuningIQA jointly learns quality regression and pairwise ranking to distinguish similar candidate images captured under different camera settings. Nevertheless, existing methods primarily output scores, attribute predictions, or preference results, without explaining the supporting visual evidence or translating residual quality issues in the preferred image into camera-parameter guidance.

\subsection{MLLM-based Image Quality Assessment}

MLLMs have extended IQA from numerical prediction to language-based quality perception and reasoning. Q-Bench, Q-Bench+, and Q-Instruct evaluate low-level perception, quality judgment, and degradation understanding. Q-Align~\cite{wu2023q}, Compare2Score~\cite{zhu2024adaptive}, and DeQA-Score~\cite{you2025teaching} estimate quality through discrete levels, quality comparisons, or score distributions. DepictQA~\cite{you2024depicting} and Co-Instruct~\cite{wu2024towards} support quality description and comparative explanation, while Q-Ground~\cite{chen2024q}, ViDA-UGC~\cite{liao2025vida}, and iDETEX~\cite{zhao2025idetex} further model localized degradations. Q-Insight~\cite{li2026q} and VisualQuality-R1~\cite{wu2026visualquality} employ reinforcement learning to improve quality scoring and reasoning.

Despite these advances, existing MLLM-based IQA methods struggle to distinguish the subtle quality differences commonly encountered in camera tuning or provide explanations grounded in such differences. Moreover, they cannot diagnose actionable issues in the preferred image or recommend corresponding camera-parameter adjustment directions. The objective of this work is to unify fine-grained quality discrimination, evidence-based explanation, and selective parameter guidance for livestreaming camera tuning.

\section{FGLive-Comments}
\subsection{Data Source}

We construct FGLive-Comments based on FGLive-10K, a fine-grained BIQA dataset for livestreaming camera tuning~\cite{sheng2026tuningiqa}. FGLive-10K contains 10,185 high-resolution images captured under diverse scenes and camera settings, with 50,925 annotations for overall quality, sharpness, noise, exposure, and face quality, as well as 19,234 fine-grained pairwise preferences. It comprises 451 training scenes and 104 test scenes. The training split contains 8,148 images and 72,678 image pairs, including 57,149 coarse-grained pairs and 15,529 fine-grained pairs, while the test split contains 2,037 images and 3,705 fine-grained pairs.

Because our task requires a clear preferred image, we discard pairs labeled as \textit{similar} and retain 13,365 fine-grained pairs. We expand their attribute annotations and pairwise preferences into structured instructions containing quality-difference explanations, global or local visual evidence, image-level summaries, and selective parameter-adjustment directions for residual issues in preferred images.

\begin{figure*}[t]
\centering
\includegraphics[width=0.85\textwidth]{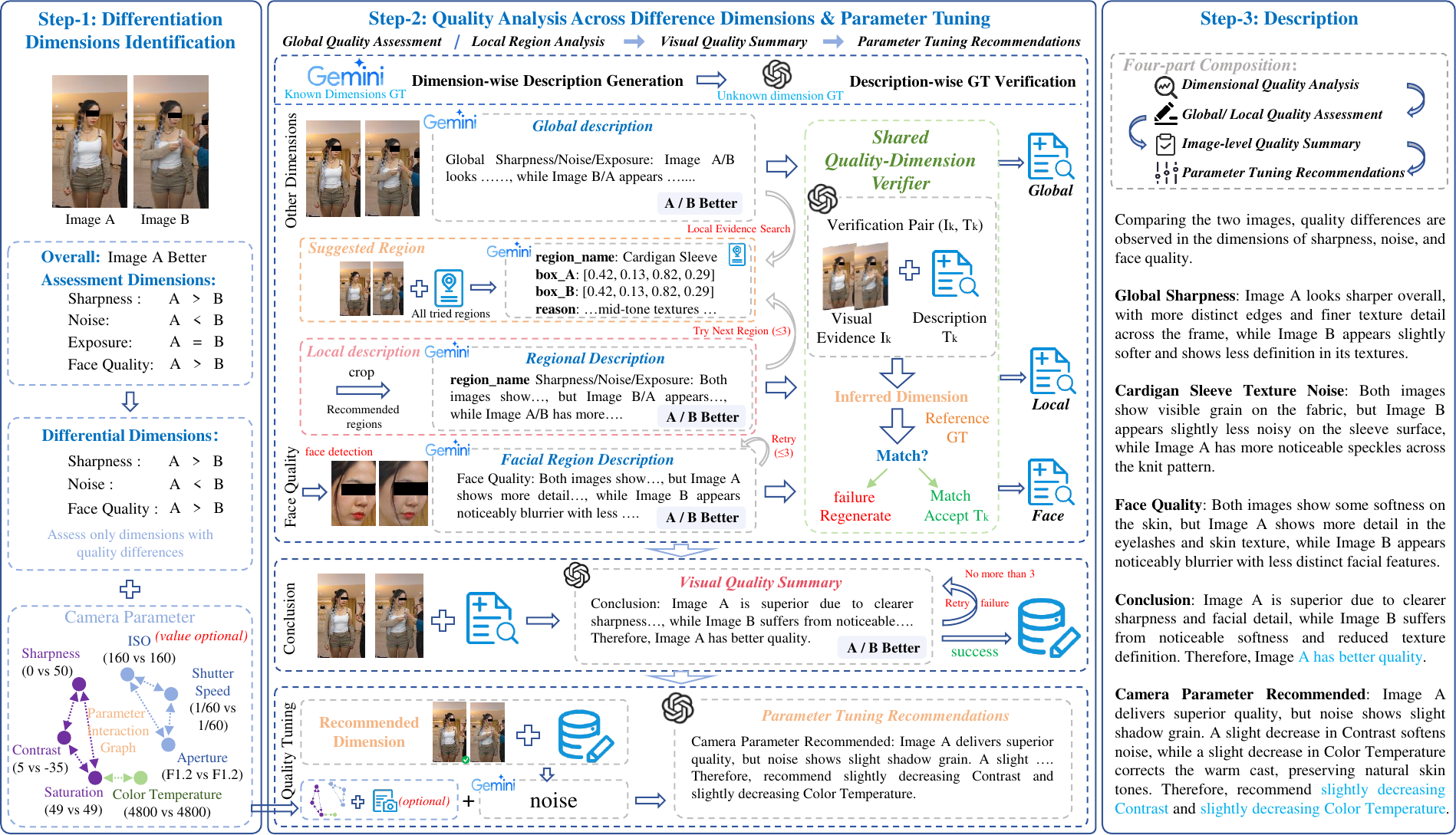} 
\caption{Dimension-wise Data Generation and Verification Pipeline.}
\label{fig:2}
\end{figure*}

\subsection{Human--Machine Collaborative Annotation}

As illustrated in Fig.~\ref{fig:2}, we construct FGLive-Comments
through a human--machine collaborative annotation pipeline. For each image pair, we determine the preferred image from the overall preference label. Among sharpness, noise, exposure, and face quality, we retain only dimensions with a clear A/B preference and exclude those labeled as \textit{similar}. Rather than directly converting these labels into text, we generate visually grounded explanations that identify quality differences and explain their contributions to the overall preference.

\textbf{Evidence-Constrained Quality Description.}
For each selected dimension, we adopt a \textit{global-to-local} evidence generation strategy. For sharpness, noise, and exposure, a global description is first generated from the image pair. A quality-dimension validator, without access to the ground-truth attribute label, then infers the relative quality from the images and description. The description is retained if its prediction matches the label; otherwise, local evidence search is performed to identify discriminative regions and generate crop-level descriptions. Each description undergoes the same verification, with up to three candidate regions examined. For face quality, facial regions are directly cropped and processed using the same generation and verification procedure. This process reduces unsupported or label-inconsistent descriptions while grounding quality explanations in discriminative visual evidence.

\begin{figure*}[t]
\centering
\includegraphics[width=0.8\textwidth]{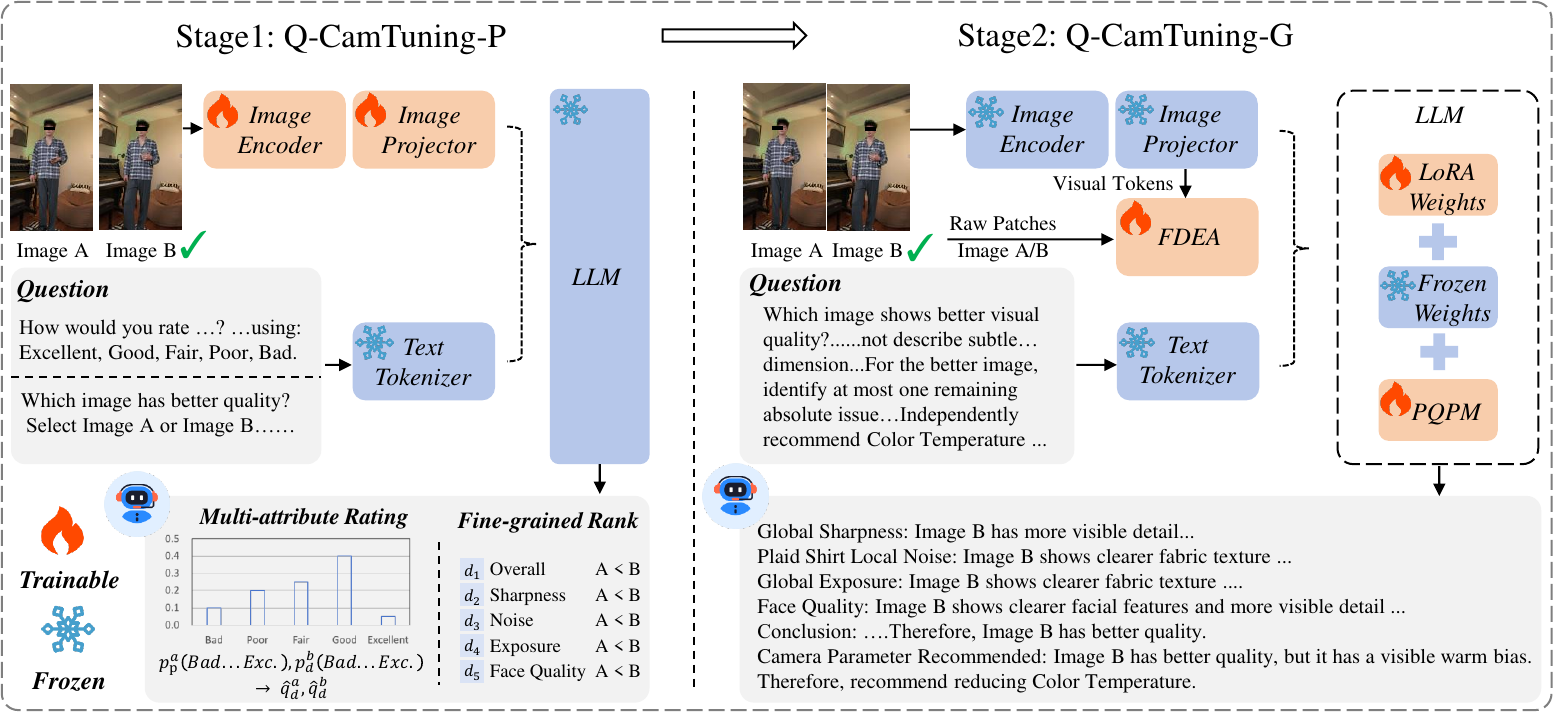} 
\caption{Progressive Training Strategy of Q-CamTuning.}
\label{fig:3}
\end{figure*}

\textbf{Image-Level Quality Summary.}
Using the validated dimension-level descriptions and their global and/or local evidence, we generate an image-level summary that integrates quality differences, identifies the main factors underlying the overall preference, and determines the preferred image. The predicted preference is extracted from the summary and checked against the FGLive-10K label. Inconsistent summaries are regenerated, ensuring alignment between the quality explanation and pairwise supervision.

\textbf{Selective Camera-Parameter Guidance.}
After determining the preferred image, we assess whether it contains residual actionable quality issues rather than continuing the pairwise comparison. A dominant issue in sharpness, noise, or exposure is identified from the validated descriptions and visual evidence, which is then mapped to the corresponding parameter-adjustment direction. Color-temperature deviations are assessed independently, with an additional direction provided for a cool or warm color cast when necessary. Thus, each sample contains at most one primary adjustment and one optional color-temperature adjustment. No recommendation is generated when the evidence is insufficient or no actionable issue exists.

Each instruction contains attribute-level analysis, global or local evidence, an image-level summary, and selective parameter guidance. After generating annotations for 13,365 training pairs, we discard samples that fail dimension-level or image-level consistency checks or lack reliable evidence after repeated attempts. The resulting training and test sets contain 12,077 and 3,291 image-description pairs, respectively. All test samples are manually verified and corrected for evidence consistency, issue validity, and compliance with the predefined parameter-mapping rules.

\section{Q-CamTuning}
As illustrated in Fig.~\ref{fig:3}, Q-CamTuning adopts a two-stage progressive training strategy that separates foundational quality perception from evidence-grounded tuning guidance. Stage-I optimizes the visual encoder and image projector while keeping the LLM frozen, using multi-attribute quality rating and fine-grained pairwise ranking supervision. The resulting model is termed Q-CamTuning-P, where ``P'' denotes \emph{perception}, as it establishes overall and attribute-level quality perception. Initialized from Q-CamTuning-P, Stage-II performs instruction tuning on FGLive-Comments to organize visual evidence into structured quality explanations, reason about image-level preferences, identify actionable issues in the preferred image, and provide selective camera-parameter recommendations. This stage produces Q-CamTuning-G, where ``G'' denotes \emph{guidance}. Such progressive training ensures that camera-parameter recommendations are grounded in reliable fine-grained quality perception.

\subsection{Progressive Perception--Evidence--Guidance Training}

\textbf{Stage-I: Perceptual Alignment.}
Stage-I focuses on visual perceptual alignment. Candidate images in livestreaming camera tuning usually share highly similar content and may differ only subtly in sharpness, noise, exposure, or face quality. To make the visual representations sensitive to such differences, we freeze the LLM while updating the visual encoder and the vision--language projection layer, thereby aligning the visual features with both overall and attribute-level quality variations.

We use three complementary supervision signals. Coarse-grained quality-distribution supervision establishes general image-level quality perception, while fine-grained pairwise ranking supervision enhances the discrimination of subtle preferences. Since fine-grained pairs contain only small quality differences and thus provide difficult ranking signals, we additionally sample coarse-grained pairs with clear quality gaps as high-confidence ranking anchors. These samples balance the difficulty of the ranking data and stabilize relative preference learning. Specifically, the quality distribution data consists of 57,149 image pairs from FGLive-10K whose overall quality score difference is larger than 0.8. Following DeQA-Score, we convert continuous quality scores into probability distributions over five quality levels: Excellent, Good, Fair, Poor, and Bad. The ranking data comprises 24,154 fine-grained pairs obtained from FGLive-Comments after removing textual descriptions and applying A/B flipping, together with 30,000 sampled coarse-grained pairs.

Let \(\mathcal{P}\) denote the training image pairs and \(\mathcal{D}=\{\mathrm{overall}, \mathrm{sharpness}, \mathrm{noise}, \mathrm{exposure}, \mathrm{face\ quality}\}\) denote the quality dimensions. For each pair \((I_a,I_b)\) and attribute \(d\), the model predicts quality distributions \(\hat{\mathbf{q}}_d^a\) and \(\hat{\mathbf{q}}_d^b\), and a binary preference probability \(\hat{c_{d}}^{ab}\). The objective of Stage-I is defined as:
\begin{equation}
\mathcal{L}_{\mathrm{S1}}
=
\lambda_{\mathrm{dist}}\mathcal{L}_{\mathrm{dist}}
+
\lambda_{\mathrm{diff}}\mathcal{L}_{\mathrm{diff}}
+
\lambda_{\mathrm{rank}}\mathcal{L}_{\mathrm{rank}} ,
\end{equation}
where \(\mathcal{L}_{\mathrm{dist}}\) aligns the predicted quality distributions with the target distributions, \(\mathcal{L}_{\mathrm{diff}}\) aligns the predicted distribution-induced score difference with the ground-truth score difference, and \(\mathcal{L}_{\mathrm{rank}}\) is a binary cross-entropy loss for pairwise preference prediction:
\begin{equation}
\mathcal{L}_{\mathrm{rank}}
=
\frac{1}{|\mathcal{P}|}
\sum_{(I_a,I_b)\in\mathcal{P}}
\frac{1}{|\mathcal{D}_{ab}|}
\sum_{d\in\mathcal{D}_{ab}}
\mathrm{BCE}
\left(
c_d^{ab},
\hat{c}_d^{ab}
\right),
\end{equation}
where \(c_d^{ab}\in\{0,1\}\) is the ground-truth preference label,
\(\hat{c}_d^{ab}\) is the predicted preference probability, and
\(\mathcal{D}_{ab}\) contains the assessment targets with valid A/B labels.

Through this stage, Q-CamTuning-P acquires overall and attribute-level quality perception, providing a stable perceptual foundation for subsequent evidence selection and quality reasoning.

\textbf{Stage-II: Evidence-Guided Preference Reasoning and Parameter Guidance.}
As shown in Fig.~\ref{fig:3}, Stage-II is initialized from Q-CamTuning-P to extend the quality perception learned in Stage-I to evidence-guided image-level preference reasoning and camera-parameter guidance. Although Q-CamTuning-P captures overall and attribute-level quality cues, different attributes depend on distinct visual evidence, which may not be adequately preserved by a shared feature representation for attribute-level descriptions. We therefore introduce the Frequency-Dimension Evidence Adapter (FDEA), which decouples evidence extraction into four attribute-aware branches to identify cues relevant to sharpness, noise, exposure, and face quality. Meanwhile, the standard autoregressive CE objective over the full response may not reliably align diagnosed quality issues with controllable camera parameters. We thus introduce the Physical-Prior Quality-to-Parameter Mapping (PQPM) module to incorporate physical quality-to-parameter priors into recommendation generation. Together, these modules enable Stage-II to organize attribute-relevant evidence into structured quality explanations and generate selective camera-parameter recommendations consistent with the diagnosed issues.

Accordingly, Stage-II adopts parameter-efficient instruction tuning on FGLive-Comments by freezing the visual encoder, vision-to-language projection layer, and base language model while jointly optimizing the LoRA, FDEA, and PQPM parameters. Given an instruction \(\mathbf{x}\) and its target response \(\mathbf{y}=(y_1,\ldots,y_T)\), the CE loss is defined as:
\begin{equation}
\mathcal{L}_{\mathrm{CE}}
=
-\frac{1}{\sum_{t=1}^{T}m_t}
\sum_{t=1}^{T}
m_t
\log p_{\theta}
\left(
y_t\mid y_{<t},I_a,I_b,\mathbf{x}
\right),
\end{equation}
where \(m_t\in\{0,1\}\) indicates whether the \(t\)-th response token is supervised. Together with the FDEA and PQPM auxiliary objectives defined below, the overall Stage-II objective is
\begin{equation}
\mathcal{L}_{\mathrm{S2}}
=
\mathcal{L}_{\mathrm{CE}}
+
\mathcal{L}_{\mathrm{FDEA}}
+
\lambda_{\mathrm{PQPM}}\mathcal{L}_{\mathrm{PQPM}},
\end{equation}
where \(\mathcal{L}_{\mathrm{FDEA}}\) denotes the combined auxiliary losses for attribute-aware evidence learning, and \(\mathcal{L}_{\mathrm{PQPM}}\) is an additional CE loss computed with a mask restricted to the parameter-recommendation field. The coefficient \(\lambda_{\mathrm{PQPM}}\) controls its contribution to the overall objective.

\begin{figure}[t]
\centering
\includegraphics[width=\columnwidth]{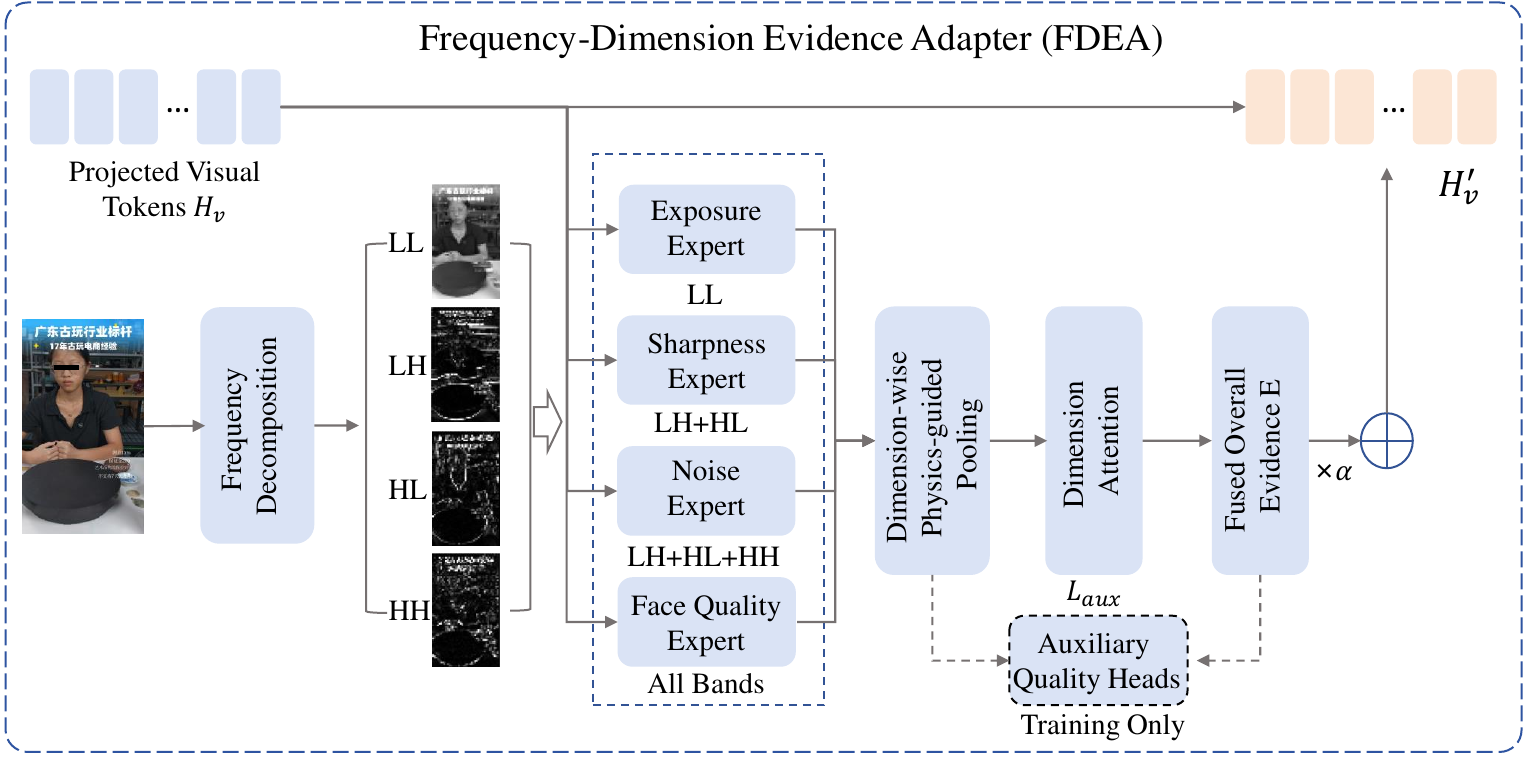} 
\caption{Frequency-Dimension Evidence Adapter.}
\label{fig:fdea}
\end{figure}

\subsection{Frequency-Dimension Evidence Adapter}

As illustrated in Fig.~\ref{fig:fdea}, FDEA applies a one-level Haar transform to raw image patches, obtaining the \(LL\), \(LH\), \(HL\), and \(HH\) frequency bands. Statistics from the selected bands are aligned with the projected visual tokens and processed by four lightweight attribute-specific branches. Specifically, the exposure branch uses \(LL\), the sharpness branch uses \(LH\) and \(HL\), the noise branch uses \(LH\), \(HL\), and \(HH\), and the face-quality branch integrates all four bands. This frequency decomposition enables each branch to extract visual evidence suited to its corresponding quality attribute.

For image \(I_i\), let \(\mathbf{H}_{v,i}\in\mathbb{R}^{N_i\times C}\) denote its \(N_i\) projected visual tokens and \(\mathbf{g}_i=\operatorname{Mean}(\mathbf{H}_{v,i})\) denote the global visual context. Given the local features \(\mathbf{Z}_i^d\) produced by attribute branch \(d\), FDEA aggregates, weights, and injects the attribute-relevant evidence as
\begin{equation}
\begin{aligned}
\mathbf{e}_i^d
&=
\operatorname{Pool}_d
(\mathbf{Z}_i^d;\mathbf{g}_i,\boldsymbol{\pi}_i^d),\\
w_i^d
&=
\operatorname{softmax}_{d}
\left(
\operatorname{Attn}(\mathbf{g}_i,\mathbf{e}_i^d)
\right),\\
\mathbf{H}_{v,i}^{\prime}
&=
\mathbf{H}_{v,i}
+
\alpha\,\mathbf{1}_{N_i}
\left(
\sum_{d\in\mathcal{A}}w_i^d\mathbf{e}_i^d
\right)^{\top},
\end{aligned}
\end{equation}
where \(\boldsymbol{\pi}_i^d\) denotes the attribute-specific frequency prior, \(\mathcal{A}\) contains the four quality attributes, and \(\mathbf{1}_{N_i}\) broadcasts the fused evidence to all visual-token positions. The coefficient \(\alpha\) controls the residual strength.

FDEA is optimized using training-only auxiliary quality heads and an attention-balancing regularizer:
\begin{equation}
\mathcal{L}_{\mathrm{FDEA}}
=
\lambda_{\mathrm{rank}}
\mathcal{L}_{\mathrm{rank}}^{\mathrm{FDEA}}
+
\lambda_{\mathrm{dist}}
\mathcal{L}_{\mathrm{dist}}^{\mathrm{FDEA}}
+
\lambda_{\mathrm{bal}}
\mathcal{L}_{\mathrm{bal}}^{\mathrm{FDEA}},
\end{equation}
where \(\mathcal{L}_{\mathrm{rank}}^{\mathrm{FDEA}}\) supervises overall and attribute-level pairwise preferences, \(\mathcal{L}_{\mathrm{dist}}^{\mathrm{FDEA}}\) aligns the predicted five-level quality distributions with their targets, and \(\mathcal{L}_{\mathrm{bal}}^{\mathrm{FDEA}}\) prevents the attribute attention from collapsing to a single branch.

\begin{figure}[t]
\centering
\includegraphics[width=\columnwidth]{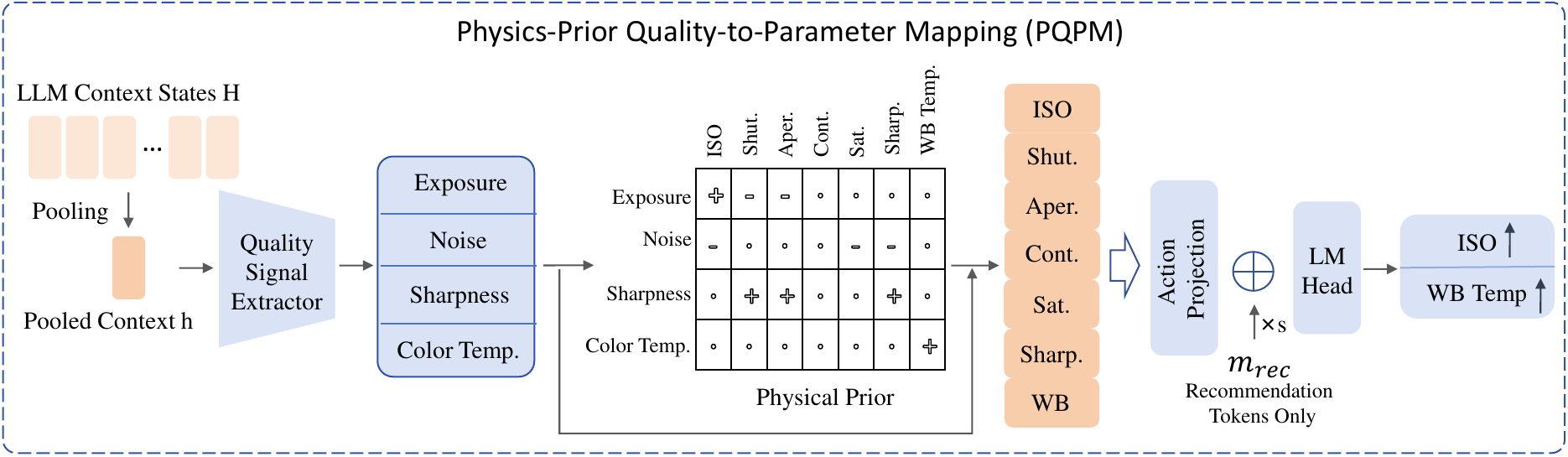}
\caption{Physical-prior quality-to-parameter mapping in PQPM.}
\label{fig:pqpm}
\end{figure}

\subsection{Physical-Prior Quality-to-Parameter Mapping}

PQPM translates the diagnosed quality issues in the preferred image into camera-parameter recommendations. As illustrated in Fig.~\ref{fig:pqpm}, it extracts quality signals for exposure, noise, sharpness, and color temperature from the language context preceding the recommendation field, and routes them to controllable camera parameters according to predefined physical relationships.

Given the context features \(\mathbf{H}^{\mathrm{ctx}}\), PQPM predicts the quality signals and maps them to a parameter-action representation:
\begin{equation}
\begin{gathered}
\mathbf{q}
=
\tanh\left(
f_q\left(\operatorname{Pool}(\mathbf{H}^{\mathrm{ctx}})\right)
\right),\\
\mathbf{a}
=
\mathbf{q}^{\top}\mathbf{M}_0,\\
\widetilde{\mathbf{h}}_t
=
\mathbf{h}_t
+
\beta m_t^{\mathrm{rec}}f_a(\mathbf{a}),
\end{gathered}
\end{equation}
where \(\mathbf{M}_0\) is a fixed physical-prior mapping matrix that encodes predefined quality-to-parameter relationships and remains unchanged during training. The projection \(f_a\) maps \(\mathbf{a}\) to the language-model hidden space, while \(m_t^{\mathrm{rec}}\) restricts its injection to the recommendation field.

\begin{table*}[t]
\centering
\setlength{\tabcolsep}{12pt}
\begin{tabular}{lcccc}
\toprule
\multirow{2}{*}{Method} & \multirow{2}{*}{Backbone} & \multicolumn{2}{c}{Score Regression} & FG Ranking \\
\cmidrule(lr){3-4} \cmidrule(lr){5-5} & 
 & SRCC$\uparrow$ & PLCC$\uparrow$ & FG-ACC$\uparrow$ \\
\midrule
DBCNN~\cite{zhang2018blind} & VGG16      & 0.751 & 0.750 & 0.567 \\
HyperIQA~\cite{su2020blindly} & ResNet-50   & 0.847 & 0.852 & 0.598 \\
MT-A~\cite{fang2020perceptual} &  ResNet-50       & 0.882 & 0.883 & 0.622 \\
CLIP-IQA+~\cite{wang2023exploring} &  CLIP (ResNet-50)  & 0.706 & 0.704 & 0.557 \\
MUSIQ~\cite{ke2021musiq} & Transformer & 0.866 & 0.869 & 0.686 \\
SARQUE~\cite{huang2022explainable} & MobileNetV3-S     & 0.858 & 0.859 & 0.610 \\
LIQE~\cite{zhang2023blind} &   CLIP (ViT-B/32)      & 0.924 & 0.921 & 0.653 \\
TuningIQA~\cite{sheng2026tuningiqa} & ResNet-50   & 0.939 & 0.936 & 0.728 \\
\midrule
Qwen3-VL-8B-Instruct~\cite{bai2025qwen3} & - & 0.807 & 0.785 & 0.540 \\
Q-Align~\cite{wu2023q} & mPLUG-Owl2-7B    & 0.872 & 0.832 & 0.631 \\
Compare2Score~\cite{zhu2024adaptive} & mPLUG-Owl2-7B & 0.850 & 0.819 & 0.627 \\
DeQA-Score~\cite{you2025teaching} & mPLUG-Owl2-7B & 0.863 & 0.864 & 0.637 \\
Q-Insight~\cite{li2026q} & Qwen-2.5-VL-7B-Instruct & 0.853 & 0.795 & 0.629 \\
VisualQuality-R1~\cite{wu2026visualquality} & Qwen-2.5-VL-7B-Instruct & 0.865 & 0.801 & 0.530 \\
\textbf{Q-CamTuning-P} & Qwen3-VL-8B-Instruct & \textbf{0.962} & \underline{0.960} & \underline{0.748} \\
\textbf{Q-CamTuning-G} & Qwen3-VL-8B-Instruct & \underline{0.961} & \textbf{0.962} & \textbf{0.756} \\
\bottomrule
\end{tabular}
\caption{Overall quality scoring and fine-grained ranking performance on FGLive-10K. The best and second-best results are highlighted in bold and underlined, respectively.}
\label{tab:overall_perception}
\end{table*}

\begin{table}[t]
\centering
\setlength{\tabcolsep}{2pt}
\begin{tabular}{llcccc}
\toprule
Method & Metric & Sharp. & Noise & Exp. & Face \\
\midrule
\multirow{3}{*}{CLIP-IQA}
& SRCC   & 0.532 & 0.472 & 0.199 & 0.510 \\
& PLCC   & 0.528 & 0.434 & 0.208 & 0.512 \\
& FG-ACC & 0.649 & 0.622 & 0.571 & 0.518 \\
\midrule
\multirow{3}{*}{MT-A}
& SRCC   & 0.896 & 0.804 & 0.782 & 0.807 \\
& PLCC   & 0.943 & 0.898 & 0.863 & 0.802 \\
& FG-ACC & 0.688 & 0.665 & 0.621 & 0.582 \\
\midrule
\multirow{3}{*}{SARQUE}
& SRCC   & 0.868 & 0.743 & 0.700 & 0.754 \\
& PLCC   & 0.930 & 0.876 & 0.804 & 0.754 \\
& FG-ACC & 0.666 & 0.645 & 0.602 & 0.647 \\
\midrule
\multirow{3}{*}{TuningIQA}
& SRCC   & \underline{0.935} & \underline{0.805} & \underline{0.897} & \underline{0.885} \\
& PLCC   & \textbf{0.963} & \underline{0.905} & \underline{0.949} & \underline{0.880} \\
& FG-ACC & \underline{0.773} & \textbf{0.746} & \underline{0.776} & \underline{0.725} \\
\midrule
\multirow{3}{*}{Q-CamTuning-P}
& SRCC   & \textbf{0.943} & \textbf{0.829} & \textbf{0.920} & \textbf{0.917} \\
& PLCC   & \underline{0.962} & \textbf{0.919} & \textbf{0.956} & \textbf{0.910} \\
& FG-ACC & \textbf{0.795} & \underline{0.744} & \textbf{0.783} & \textbf{0.747} \\
\bottomrule
\end{tabular}
\caption{Attribute-level quality scoring and ranking performance on FGLive-10K. Highlighting follows Table~\ref{tab:overall_perception}.}
\label{tab:attribute_perception}
\end{table}

\section{Experiments}
\subsection{Experimental Setup}

\textbf{Datasets and Metrics.}
We evaluate Q-CamTuning on FGLive-10K and FGLive-Comments using the scene-level split of FGLive-10K. We report SRCC and PLCC for quality scoring and ACC for overall and attribute-level pairwise ranking. Following DepictQA~\cite{you2024depicting}, GPT-5.5 evaluates quality descriptions against ground-truth descriptions to compute GPT-Score. Parameter recommendations is assessed through a blind subjective study using preference rate and acceptance score, which measure wins comparable cases, respectively.

\textbf{Task-Specific Evaluation Sets.}
For preferred-image issue identification, we construct the \textit{\textbf{Preferred-Image Issue Evaluation Set}} from preferred images in the FGLive-Comments test split. Each image is assigned a primary quality-issue label from sharpness, noise, exposure, and no actionable issue. Since an image may additionally exhibit a color-temperature deviation, the presence of a color-temperature issue is annotated independently. Labels are verified through GPT-assisted human review: GPT-consistent samples are checked by one annotator, while ambiguous or inconsistent samples are independently reviewed by three annotators with majority voting. The resulting set contains 1,058 images.

For parameter-adjustment recommendation, we construct the \textit{\textbf{Parameter-Guidance Evaluation Set}} by sampling 200 test pairs based on the verified issue labels. The set is balanced across sharpness, noise, exposure, and no-actionable-issue cases, with color-temperature cases accounting for approximately one quarter.

\textbf{Implementation Details.}
Q-CamTuning is initialized from Qwen3-VL-8B~\cite{bai2025qwen3}, which also serves as the baseline. Stage-I trains the visual encoder and projection layer with learning rates of \(1\times10^{-5}\) and \(5\times10^{-5}\), respectively. Stage-II freezes the visual encoder, retains the trainable projection layer, and applies LoRA~\cite{hu2022lora} to the language model with a learning rate of \(1\times10^{-5}\). Training uses AdamW with cosine annealing, a batch size of 128, and one epoch on two NVIDIA RTX PRO 6000 GPUs.

\subsection{Evaluation of Quality Perception}

\textbf{Overall Quality Assessment.}
Table~\ref{tab:overall_perception} compares Q-CamTuning with existing BIQA methods, the task-specific TuningIQA model, and recent IQA-oriented MLLMs on FGLive-10K. Q-CamTuning-P achieves the best SRCC and PLCC, showing that Stage-I perceptual alignment effectively establishes strong image-level quality scoring ability. For overall fine-grained ranking, Q-CamTuning-G further improves FG-ACC after evidence-guided instruction tuning, demonstrating that selecting and organizing global--local evidence helps image-level preference reasoning in camera-tuning scenarios.

\textbf{Attribute-Level Quality Assessment.}
To evaluate attribute-aware quality perception, we compare Q-CamTuning-P with three SOTA methods that support quality attribute prediction: CLIP-IQA, MT-A, SARQUE, and TuningIQA. As shown in Table~\ref{tab:attribute_perception}, Q-CamTuning-P achieves competitive performance across all quality attributes, providing a solid perceptual foundation for attribute-level evidence selection in Stage-II.

\begin{table}[t]
\setlength{\tabcolsep}{1pt}
\centering
\begin{tabular}{lcc}
\toprule
Method
& Primary Quality Issue $\uparrow$ 
& Color Temp. $\uparrow$ \\
\midrule
Qwen3-VL-8B 
&  0.436 & 0.656 \\
Q-Insight & 0.423 & \underline{0.657} \\
GPT-5.4 & \underline{0.595} & 0.614 \\
Q-CamTuning-G & \textbf{0.682} & \textbf{0.700} \\
\bottomrule
\end{tabular}
\caption{Accuracy of dominant quality-issue identification. Highlighting follows Table~\ref{tab:overall_perception}.}
\label{tab:preferred_issue}
\end{table}

\textbf{Preferred-Image Adjustable Issue Identification.}
On the Preferred-Image Issue Evaluation Set, we compare Q-CamTuning-G with description-capable MLLMs, including Qwen3-VL-8B, Q-Insight, and GPT-5.4. As shown in Table~\ref{tab:preferred_issue}, Q-CamTuning-G substantially outperforms the open-source baselines in dominant-issue and color-temperature identification while remaining competitive with GPT-5.4, validating the effectiveness of evidence-guided training.

\begin{table}[t]
\centering
\setlength{\tabcolsep}{10pt}
\begin{tabular}{lc}
\toprule
Method & GPT-Score$\uparrow$ \\
\midrule
Qwen3-VL-8B-Instruct & 0.421 \\
Q-Insight & 0.247 \\
GPT-5.4 & 0.515 \\
\midrule
Q-CamTuning-P & 0.527 \\
Q-CamTuning-P w/ Stage-II & \underline{0.575} \\
Q-CamTuning-P w/ Stage-II + FDEA & \textbf{0.599} \\
\bottomrule
\end{tabular}
\caption{GPT-Score results for quality description evaluation. Highlighting follows Table~\ref{tab:overall_perception}.}
\label{tab:description_eval}
\end{table}

\subsection{Evaluation of Quality Description and Parameter Guidance}

\textbf{Quality Description Evaluation.}
Table~\ref{tab:description_eval} reports GPT-Score, which evaluates generated quality descriptions against ground-truth references. Q-CamTuning-G achieves higher scores than existing MLLMs. Its improvements over Q-CamTuning-P and the variant without FDEA demonstrate the effectiveness of Stage-II and FDEA, respectively, in evidence-grounded quality description.

\begin{table}[t]
\centering
\setlength{\tabcolsep}{15pt}
\begin{tabular}{lc}
\toprule
Method & Pref.$\uparrow$ \\
\midrule
Q-CamTuning-G vs. Qwen3-VL-8B & 0.610 \\
Q-CamTuning-G vs. Q-Insight & 0.643\\
Q-CamTuning-G vs. GPT-5.4 & 0.543 \\
Q-CamTuning-G  vs. w.o./ PQPM  & 0.580 \\
\bottomrule
\end{tabular}
\caption{Subjective evaluation of parameter-direction recommendation. Pref. denote preference rate.}
\label{tab:guidance_subjective}
\end{table}

\textbf{Subjective Evaluation of Parameter Guidance.}
We conduct a blind subjective study on the Parameter-Guidance Evaluation Set. Without access to issue labels, color-temperature labels, or method identities, three reviewers compare anonymized recommendations from Q-CamTuning-G and each baseline. Each sample is classified by majority vote as favoring either method, comparable, or neither reasonable. As shown in Table~\ref{tab:guidance_subjective}, Q-CamTuning-G achieves higher preference rates, demonstrating its parameter-recommendation capability. Its improvement over the variant without PQPM further confirms that PQPM enhances the reasonableness of the recommendations.

\begin{figure}[!h]
\centering
\includegraphics[width=\columnwidth]{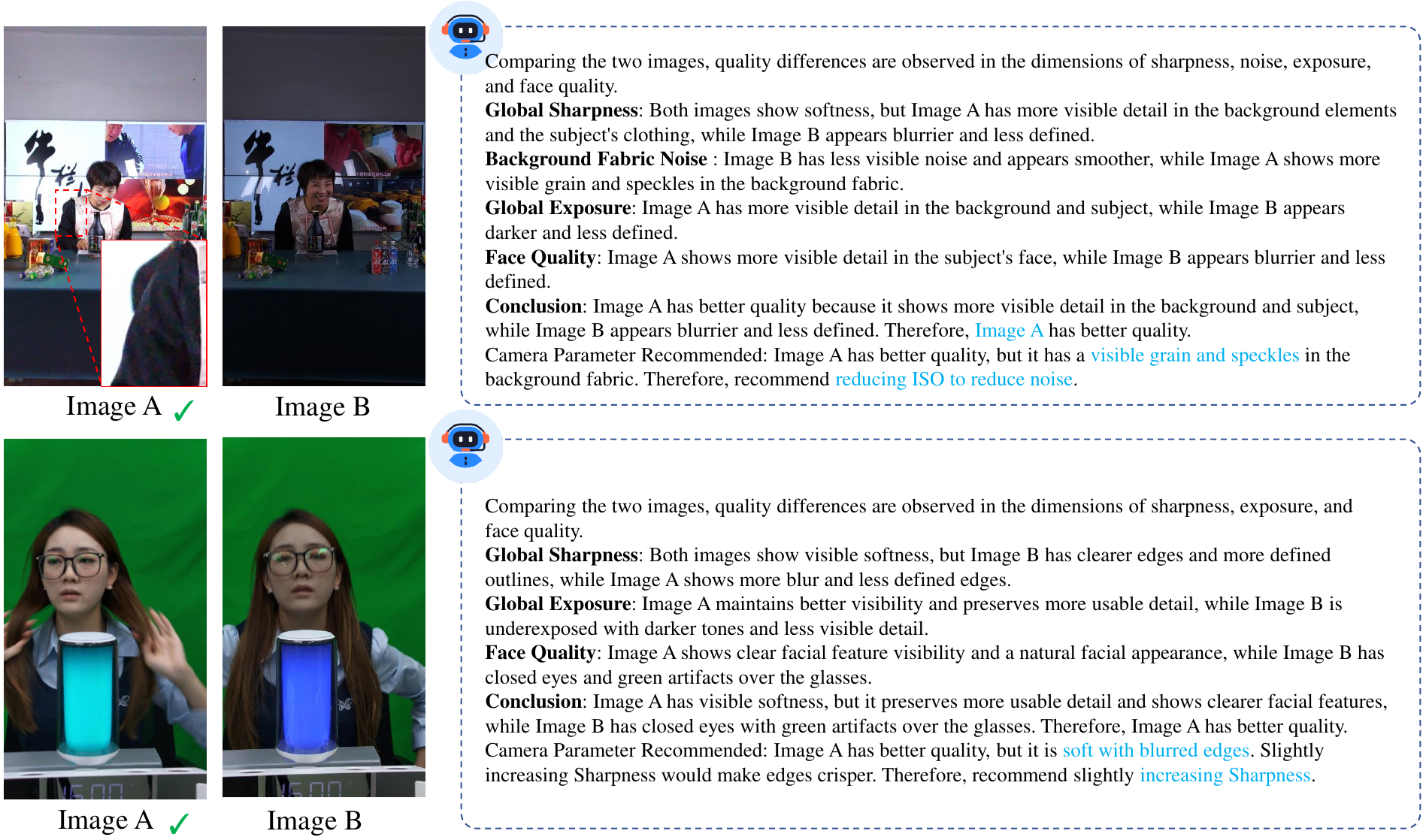}
\caption{Qualitative examples of evidence-grounded quality descriptions and camera-parameter guidance. Local evidence is highlighted in red.}
\label{fig:parameter_case}
\end{figure}

\subsection{Qualitative Analysis}

Figure~\ref{fig:parameter_case} presents qualitative results generated by Q-CamTuning-G. The model identifies subtle attribute-level differences using global and local visual evidence, supports its overall preference judgment, and provides actionable camera-parameter recommendations for issues in the preferred image.

\section{Conclusion}

This paper presents Q-CamTuning, an MLLM for livestreaming camera tuning that unifies fine-grained quality assessment, evidence-grounded reasoning, and camera-parameter guidance. We construct FGLive-Comments with global--local visual evidence and selective tuning recommendations, and develop a progressive perception--evidence--guidance framework enhanced by frequency-aware local evidence and quality-to-parameter mapping. Experiments demonstrate strong performance in quality scoring and fine-grained ranking, together with evidence-grounded descriptions and reasonable parameter recommendations. Future work will explore precise parameter-value prediction and real-world closed-loop camera tuning.

\bibliography{aaai2027}


\end{document}